%% file: main.tex
\documentclass[sigconf,nonacm]{acmart}

\usepackage[caption=false]{subfig}
\usepackage{multirow}

\AtBeginDocument{%
  \providecommand\BibTeX{{%
    \normalfont B\kern-0.5em{\scshape i\kern-0.25em b}\kern-0.8em\TeX}}}

\begin{document}

\author{Wenkai Li}
\affiliation{
\institution{Tsinghua University}
\city{Beijing}
\country{China}
}
\email{liwk20@mails.tsinghua.edu.cn}

\author{Cheng Feng}
\affiliation{%
  \institution{Siemens AG}
  \city{Beijing}
  \country{China}
}
\email{cheng.feng@siemens.com}

\author{Ting Chen}
\affiliation{%
  \institution{Tsinghua University}
  \city{Beijing}
  \country{China}}
\email{tingchen@tsinghua.edu.cn}

\author{Jun Zhu}
\affiliation{%
  \institution{Tsinghua University}
  \city{Beijing}
  \country{China}}
\email{dcszj@tsinghua.edu.cn}

\begin{abstract}
Time series anomaly detection (TSAD) is an important data mining task
with numerous applications in the IoT era. In recent years, a large number of deep neural network-based methods have been proposed, demonstrating significantly better performance than conventional methods on addressing challenging TSAD problems in a variety of areas. Nevertheless, these deep TSAD methods typically rely on a clean training dataset that is not polluted by anomalies to learn the "normal profile" of the underlying dynamics. This requirement is nontrivial since a clean dataset can hardly be provided in practice. Moreover, without the awareness of their robustness, blindly applying deep TSAD methods with potentially contaminated training data can possibly incur significant performance degradation in the detection phase. In this work, to tackle this important challenge, we firstly investigate the robustness of commonly used deep TSAD methods with contaminated training data which provides a guideline for applying these methods when the provided training data are not guaranteed to be anomaly-free. Furthermore, we propose a model-agnostic method which can effectively improve the robustness of learning mainstream deep TSAD models with potentially contaminated data. Experiment results show that our method can consistently prevent or mitigate performance degradation of mainstream 
deep TSAD models on widely used benchmark datasets.
\end{abstract}

\title{Robust Learning of Deep Time Series Anomaly Detection Models with Contaminated Training Data}

\maketitle

\section{Introduction}
Recent advances in sensor technology allow us to collect a large amount of time series data in a variety of areas. With numerous applications in intrusion detection, system health monitoring and predictive maintenance, time series anomaly detection (TSAD) which aims to find temporal signals that deviate significantly from other observations becomes an important data mining task. With growing complexity of collected data, conventional TSAD models such as dynamical state space models \cite{ding2020secure}, autoregressive models \cite{gunnemann2014robust,melnyk2016vector}, rule-based models \cite{feng2019systematic,dhaou2021causal} generally suffer from sub-optimal performance due to their limited capacity to capture nonlinear system dynamics in high dimensional space. 

In recent years, deep neural networks (DNNs) are widely used, demonstrating significantly better performance than conventional anomaly detection models on addressing challenging TSAD problems in a variety of real-world applications. To date, the mainstream deep TSAD models can be categorized into two types: prediction-based and reconstruction-based. Specifically, prediction-based TSAD first learns a predictive model such as recurrent neural networks \cite{hundman2018detecting,tariq2019detecting,feng2017multi,wu2020developing} and convolutional neural networks \cite{wen2019time,he2019temporal} to predict signals for future time steps. Then the predicted signals are compared with observed ones to generate a residual error. An anomaly is detected if the residual error exceeds a threshold. Reconstruction-based TSAD takes a similar approach except that it learns a reconstruction model such as autoencoders \cite{audibert2020usad,kieu2019outlier,malhotra2016lstm,park2018multimodal,zhang2019deep} and transformers \cite{tuli2022tranad,meng2019spacecraft} to compress temporal signals to lower dimensional embeddings and reconstruct them afterwards. The reconstruction error is used to detect anomalies. There are also other deep anomaly detection models such as density-based \cite{an2015variational,su2019robust,zong2018deep,xu2018unsupervised,feng2021time,li2021multivariate}, generative adversarial network-based \cite{schlegl2017unsupervised,schlegl2019f,li2018anomaly,li2019mad} and one class-based \cite{ruff2018deep,wu2019deep}. However, they are either not specifically designed for time series data or closely related to the mainstream models since their underlying backbone is mostly a reconstruction or prediction model.

Due to scarcity of labelled anomalies in general, most deep TSAD models are semi-supervised, meaning that they require a clean training dataset that is not polluted by any anomalies to learn the normal profile of the temporal dynamics within the data. However, such a clean dataset rarely exists in practice. In real applications, the modellers are often provided with datasets that are likely to be polluted with unknown anomalies. Meanwhile, DNNs are often trained with an over-parameterized regime, meaning that the number of their parameters can exceed the size of the training data. As a result, deep TSAD models have the capacity to overfit to any given training data regardless of the unknown ratio of anomalies, leading to serious performance degradation in the detection phase. This problem is clearly demonstrated in our experiments using two representative deep TSAD models (a prediction-based and a reconstruction-based) on various benchmark datasets.

Furthermore, we propose a simple yet effective model-agnostic method which can significantly improve the robustness of deep TSAD models learned from contaminated training data. Specifically, our method is to discard samples with consistently large training losses or strong oscillation on loss updates during the early phase of model training. Our sample filtering method is justified by the "memorization effect" \cite{arpit2017closer} of DNNs, which means that DNNs tend to prioritize
learning simple patterns that are shared by multiple samples first and then gradually memorize noisy samples during training. In our experiments, we show that our method can consistently prevent or mitigate performance degradation of mainstream deep TSAD
models on various benchmark datasets.

\section{Methodology}
Our model-agnostic sample filtering method for learning robust deep TSAD models is based on the observation that DNNs start to learn from common patterns in initial phases and gradually adapt to noisy samples during training \cite{arpit2017closer}. Thus when trained on contaminated data, DNNs will prioritize learning from normal samples before over-fitting to the whole dataset. Intuitively, when DNNs update their parameters using backpropagation, the gradient is averaged over all training data. As a result, the overall gradient is more likely to be dominated by the normal samples during the initial phase. For this reason, the training losses on normal samples are more likely to decrease steadily, but stay large or oscillate strongly on abnormal samples over the training epochs in the initial phase. 

\subsection{Filtering Plausible Abnormal Samples}
Based on the above observation, we propose two metrics by which we filter out anomalous training samples using a few trial epochs. Specifically, let $N$ be the total number of trial epochs and $\mathbf{x}$ be a training datum, we use $L_{\mathbf{x}}^i$ to denote the loss of ${\mathbf{x}}$ at the end of $i$th epoch and $\Delta_{\mathbf{x}}^i = L_{\mathbf{x}}^i-L_{\mathbf{x}}^{i-1}$ to denote the update of loss at the $i$th epoch. The following two metrics are calculated for each datum at the end of the $N$th trial epoch:
\begin{eqnarray}
m_{\mathbf{x}} &=& \mu(L_\mathbf{x})  \quad \text{where}\ L_\mathbf{x} = [L_\mathbf{x}^1, \dots, L_\mathbf{x}^N] \\
v_{\mathbf{x}} &=& \sigma(\Delta_\mathbf{x}) \quad \text{where}\ \Delta_\mathbf{x} = [\Delta_\mathbf{x}^1, \dots, \Delta_\mathbf{x}^N]
\end{eqnarray}
where $m_{\mathbf{x}}$ is the mean of losses for $\mathbf{x}$ during the trial epochs, $v_{\mathbf{x}}$ is the standard deviation of loss updates for $\mathbf{x}$ during the trial epochs. Concretely, $m_{\mathbf{x}}$ is used to filter out samples with larger training losses; $v_{\mathbf{x}}$ is used to filter out samples with stronger oscillations on per-epoch loss updates indicating that the gradient direction of the datum is inconsistent with normal samples. A higher value of either metric indicates the sample is likely to be an anomaly.

\subsection{Discarding Plausible Abnormal Samples}
\label{sec:discard}
Let $\tau$ to be the user-defined upper bound of anomaly ratio in the potentially contaminated dataset, we let $S_m=\{ \mathbf{x} \mid m_{\mathbf{x}}>Q_m(1-\tau)  \}$ and $S_v=\{ \mathbf{x} \mid v_{\mathbf{x}}>Q_v(1-\tau)  \}$ where $Q_m(1-\tau)$ and $Q_v(1-\tau)$ are the $1-\tau$ quantile of $m_{\mathbf{x}}$ and $v_{\mathbf{x}}$ in the dataset, respectively. We discard all samples in $S_m \cup S_v$. After discarding the plausible abnormal samples, we retrain the model 
on the remaining data from scratch. Our sample discarding strategy seems rather aggressive, however, as the anomaly ratio in most cases is a small value and time series data oftentimes enjoy a high degree of redundancy due to repeated patterns caused by seasonality, discarding a small portion of normal data is unlikely to cause performance degradation of the trained model.

\section{Experiments}
The goal of our experiments is two-fold. Firstly, we investigate the robustness of various deep TSAD models with contaminated training data. Secondly, we show whether our proposed method can effectively improve the robustness of those deep TSAD models.  


\subsection{Benchmark Deep TSAD Models}
We first define the models that we will analyze. Since it is impossible to cover all deep TSAD models, we first select one model in the reconstruction-based class and another in the prediction-based class as the representative mainstream deep TSAD models. To make our experiments more comprehensive, we also select one density-based model (DAGMM \cite{zong2018deep}) and one one-class classification deep anomaly detection model (DeepSVDD~\cite{ruff2018deep}). Although these two models are not designed for time series data, they are frequently used as baselines in the deep TSAD literature. More details of selected models are given as follows:
\begin{itemize}
\item LSTMAE: LSTMAE is a recurrent autoencoder implemented by Long Short Term Memory (LSTM) \cite{hochreiter1997long} networks. It is the "vanilla" version of reconstruction-based deep TSAD model that has been the backbone of many more complicated models \cite{kieu2019outlier,malhotra2016lstm,park2018multimodal} in this class.
 \item Seq2SeqPred: Seq2SeqPred is a prediction-based model implemented as a sequence to sequence LSTM \cite{sutskever2014sequence}. It is a commonly used deep time series prediction model for anomaly detection.
\item DAGMM: DAGMM is a density-based anomaly detection model using a deep generative model that assumes the Gaussian mixture prior in the latent space to estimate the likelihood of input samples. It is commonly used as a baseline model in the deep TSAD literature.
\item DeepSVDD: DeepSVDD is an one-class classification deep anomaly detection model that is also commonly used as a baseline in the deep TSAD literature .
\end{itemize}

\subsection{Benchmark Datasets}
We consider the following four commonly used benchmark datasets for TSAD:
\begin{itemize}
    \item SWAT~\cite{swat}: This dataset is collected from a testbed which is a scaled down version of a real-world industrial water treatment plant producing filtered water. It consists of operation records collected per second for 11 days with the first 7 days running normally.  
    \item WADI~\cite{wadi}: This dataset is collected from a testbed which represents a scaled down version of an urban water distribution system. The dataset contains records of each second in 16 days with the former 14 days running normally. The data in the last day is ignored as they have different distributions with previous 15 days due to change of operational mode.
    \item PUMP~\cite{feng2021time}: This dataset is collected from a water pump system of a small town. It contains data points of every minute in 5 months.
    \item PSM~\cite{psm}: This dataset, collected by eBay, consists of recorded server machine metrics per minute for 21 weeks, in which 13 weeks are for training and 8 weeks are for testing. 
\end{itemize}
All the above datasets consist of a training set which is anomaly-free and a test set which contains timely distributed anomalies. To conduct our experiments with contaminated training data, we pollute these training sets by randomly injecting anomalous windows sampled from test data. Specifically, we pollute training sets at different anomaly ratios from 0\% to 20\%. The performance at the ratio 0\% tells the performance of models when using clean training data. As for the range upper bound, it is sufficiently high since anomaly ratios of real-world datasets seldomly exceed 15\%. Concretely, we select $[0\%, 1\%, 2\%, 3\%, 4\%, 6\%, 8\%, 10\%, 13\%, 16\%, 20\%]$, 11 anomaly ratios in total.

\subsection{Evaluation Metrics}
To date, AUC-ROC score, the best F1 score and the best point-adjusted F1 score are the most commonly adopted evaluation metrics for TSAD. However, \cite{kim2021towards} reveals that the best point-adjusted F1 score has a great possibility of overestimating the detection performance and even a random anomaly score can easily turn into a state-of-the-art TSAD method. Thus we use AUC-ROC score and the best F1 score as our evaluation metrics. Specifically, for the best F1 score, it is reported as the best F1 score by searching all possible anomaly thresholds.

\subsection{Baseline Training Methods}
To demonstrate the benefits of our proposed sample filtering method in improving the robustness of trained deep TSAD models with contaminated data, we compare our method with the following three baseline training methods:
\begin{itemize}
\item Vanilla method: We just train the models with the contaminated data without any adjustment. The objective of comparing with this baseline is two-fold: investigating the robustness of the benchmark models and show the superiority of our proposed method.
\item $m_{\mathbf{x}}$ filtering method: We only discard samples with larger mean training losses in the trial epochs. That is to say, we only discard samples in $S_m$ as defined in Section~\ref{sec:discard}.
\item $v_{\mathbf{x}}$ filtering method: We only discard samples with larger oscillations on per-epoch loss updates in the trial epochs. That is to say, we only discard samples in $S_v$ as defined in Section~\ref{sec:discard}.
\end{itemize}

\subsection{Experiment Results}



We have 4 benchmark deep TSAD models, 4 benchmark datasets and 4 training methods for each model. To produce convincing results, each benchmark model under a certain training method is trained on each training set over 11 selected contamination ratios with 5 times' repetition, and then we report the average anomaly detection performance on the test sets with standard deviation. For all experiments except the vanilla method, the value of $\tau$ (the upper bound of anomaly ratio) is fixed at $0.2$, $N$ (the number of trial epochs) is fixed at $10$. 

Regarding model hyperparameters, the length of reconstruction and prediction window for LSTMAE and Seq2SeqPred is fixed at 12 for SWAT and WADI, 5 for PUMP and 20 for PSM respectively. After discarding plausible abnormal samples at the end of the trial epochs, we split the filtered training data according to the ratio 4:1, among which 1 out of 5 data is used for validation. Randomised grid search is used to tune other model-specific hyperparameters for best validation loss.  



\subsubsection{Results for Reconstruction-based and Prediction-based Models}
\input{table}
\input{pic_summary/lstmaepic}

\input{pic_summary/seqpic}
The performance of LSTMAE and Seq2SeqPred with contaminated training data under different anomaly ratios are given in Fig.~\ref{fig:lstmae} and Fig.~\ref{fig:seq} respectively. Notably, the AUC and best-F1 curves for the vanilla method decrease sharply as the contamination ratio increasing from 0\%, indicating the poor robustness of the original LSTMAE and Seq2SeqLSTM models with contaminated training data. \emph{As a result, blindly applying these deep TSAD models with potentially contaminated training data has a great possibility of encountering unexpected serious performance degradation in the detection phase.}

Furthermore, after applying the filtering methods, we find that the robustness of both models can be significantly improved. Specifically, the AUC and best-F1 curves of $m_{\mathbf{x}}$ filtering, $v_{\mathbf{x}}$ filtering and our method all lie above the one from the vanilla method by a clear margin. The comparison between $m_{\mathbf{x}}$ filtering and $v_{\mathbf{x}}$ filtering empirically justify our motivation when designing these two metrics. $v_{\mathbf{x}}$ filtering achieves better performance on SWAT and PUMP while $m_{\mathbf{x}}$ filtering performs better in PSM. In Tab.~\ref{tab:coverage}, we show the coverage (avg. $\pm$ standard deviation) of injected anomalies by the discarded samples of different filtering methods under 6\%, 13\% and 20\% contamination ratios. It can be seen that using $v_{\mathbf{x}}$ can filter more abnormal samples than $m_{\mathbf{x}}$ in SWAT, WADI and PUMP, whilst $m_{\mathbf{x}}$ in many cases filter out more abnormal samples in PSM. \emph{By leveraging the advantages of both metrics, our method constantly achieves the best performance in all scenarios}. 

It is worthy to note that even discarding more than $20\%$ of data, our method achieves a similar performance with the vanilla method under zero contamination ratio. \emph{This means that dropping a proportion of data by our method will be unlikely to damage the performance of original reconstruction-based or prediction-based deep TSAD model with a much lower contamination ratio}.

\subsubsection{Results for Density-based and One-Class Classification Models}
\input{pic_summary/dagmmpic}

\input{pic_summary/deepsvddpic}
As illustrated in Fig.~\ref{fig:dagmm}, compared with LSTMAE and Seq2SeqPred, DAGMM is not so sensitive to contaminated training data. This is because DAGMM assumes that lower embeddings of training samples are governed by a mixture of Gaussian distributions, which is a strong prior that limits the capacity of the model and further prevents over-fitting to abnormal samples. DAGMM tends to underfit the samples because of its limited model capacity. Thus, we believe that when the model underfits, the loss traces are no longer reliable to distinguish the abnormal samples and our sample filtering method does not work in these cases. 

Instead of learning the distribution of normal samples, DeepSVDD aims to squeeze all training samples into a hypersphere whose radius is as small as possible. Since all samples, no matter normal ones or abnormal ones, are pushed to the center point, their loss traces smoothly approach 0 which are uninformative. Thus our sample filtering method also does not work for one-class classification models as illustrated in Fig.~\ref{fig:deepsvdd}.

\section{Discussion and Conclusion}
We study the robustness of deep TSAD models with contaminated training data, a problem of significant practical meaning. We show that the mainstream deep TSAD models have poor robustness to contaminated training data. Furthermore, we show that our proposed sample filtering method can effectively prevent or mitigate performance degradation of mainstream deep TSAD models under contaminated training data.

Robustness of anomaly detection models with contaminated training data have been studied before. For example, in \cite{zhang2021elite}, a method called ELITE is proposed which uses a small number of labeled anomalies to infer the anomalies hidden in the training samples. Unlike ELITE, our method does not require any labeled samples to filter out anomalies. More similar work to ours are \cite{xia2015learning} and \cite{du2021gan} where samples with larger losses are filtered out during model training to enhance the robustness of anomaly detection models. In this work, we show that filtering out plausible abnormal samples with stronger oscillations on per-epoch loss updates in the initial phase of model training can further enhance the robustness of deep TSAD models significantly.

\bibliographystyle{ACM-Reference-Format}
\bibliography{ref}

\appendix

\end{document}

%% file: table.tex
\begin{table*}[htbp]
\caption{Coverage of injected anomalies by the discarded samples under 6\%, 13\% and 20\% contamination ratios}
\label{tab:coverage}
\begin{tabular}{cccccccccccccc}
\hline
\multirow{2}{*}{Model}                                                   & \multirow{2}{*}{Filter} & \multicolumn{3}{c}{SWAT}                                                                                                                                                                      & \multicolumn{3}{c}{WADI}                                                                                                                                                                       & \multicolumn{3}{c}{PUMP}                                                                                                                                                                         & \multicolumn{3}{c}{PSM}                                                                                                                                                  \\
                                                                         &                         & 6\%                                                    & 13\%                                                   & \multicolumn{1}{c|}{20\%}                                                   & 6\%                                                     & 13\%                                                   & \multicolumn{1}{c|}{20\%}                                                   & 6\%                                                     & 13\%                                                    & \multicolumn{1}{c|}{20\%}                                                    & 6\%                                                    & 13\%                                                   & 20\%                                                   \\ \hline
\multirow{3}{*}{LSTMAE}                                                  & $m_{\mathbf{x}}$ filtering                    & \begin{tabular}[c]{@{}c@{}}96.33\\ (0.46)\end{tabular} & \begin{tabular}[c]{@{}c@{}}61.83\\ (7.83)\end{tabular} & \multicolumn{1}{c|}{\begin{tabular}[c]{@{}c@{}}43.38\\ (1.30)\end{tabular}} & \begin{tabular}[c]{@{}c@{}}41.57\\ (2.15)\end{tabular}  & \begin{tabular}[c]{@{}c@{}}27.38\\ (3.21)\end{tabular} & \multicolumn{1}{c|}{\begin{tabular}[c]{@{}c@{}}20.88\\ (1.13)\end{tabular}} & \begin{tabular}[c]{@{}c@{}}100.00\\ (0.00)\end{tabular} & \begin{tabular}[c]{@{}c@{}}96.79\\ (3.12)\end{tabular}  & \multicolumn{1}{c|}{\begin{tabular}[c]{@{}c@{}}71.13\\ (1.46)\end{tabular}}  & \begin{tabular}[c]{@{}c@{}}90.41\\ (5.16)\end{tabular} & \begin{tabular}[c]{@{}c@{}}81.09\\ (4.52)\end{tabular} & \begin{tabular}[c]{@{}c@{}}69.22\\ (5.22)\end{tabular} \\
                                                                         & $v_{\mathbf{x}}$ filtering                     & \begin{tabular}[c]{@{}c@{}}89.12\\ (2.44)\end{tabular} & \begin{tabular}[c]{@{}c@{}}88.39\\ (1.21)\end{tabular} & \multicolumn{1}{c|}{\begin{tabular}[c]{@{}c@{}}84.65\\ (5.74)\end{tabular}} & \begin{tabular}[c]{@{}c@{}}78.86\\ (7.48)\end{tabular}  & \begin{tabular}[c]{@{}c@{}}84.10\\ (2.85)\end{tabular} & \multicolumn{1}{c|}{\begin{tabular}[c]{@{}c@{}}86.59\\ (2.10)\end{tabular}} & \begin{tabular}[c]{@{}c@{}}99.84\\ (0.17)\end{tabular}  & \begin{tabular}[c]{@{}c@{}}99.83\\ (0.13)\end{tabular}  & \multicolumn{1}{c|}{\begin{tabular}[c]{@{}c@{}}99.68\\ (0.20)\end{tabular}}  & \begin{tabular}[c]{@{}c@{}}68.84\\ (5.35)\end{tabular} & \begin{tabular}[c]{@{}c@{}}73.10\\ (4.13)\end{tabular} & \begin{tabular}[c]{@{}c@{}}68.99\\ (7.53)\end{tabular} \\
                                                                         & ours                   & \begin{tabular}[c]{@{}c@{}}97.90\\ (0.92)\end{tabular} & \begin{tabular}[c]{@{}c@{}}95.88\\ (1.09)\end{tabular} & \multicolumn{1}{c|}{\begin{tabular}[c]{@{}c@{}}94.16\\ (5.57)\end{tabular}} & \begin{tabular}[c]{@{}c@{}}82.11\\ (4.70)\end{tabular}  & \begin{tabular}[c]{@{}c@{}}84.49\\ (2.84)\end{tabular} & \multicolumn{1}{c|}{\begin{tabular}[c]{@{}c@{}}87.80\\ (1.79)\end{tabular}} & \begin{tabular}[c]{@{}c@{}}100.00\\ (0.00)\end{tabular} & \begin{tabular}[c]{@{}c@{}}100.00\\ (0.00)\end{tabular} & \multicolumn{1}{c|}{\begin{tabular}[c]{@{}c@{}}100.00\\ (0.00)\end{tabular}} & \begin{tabular}[c]{@{}c@{}}96.25\\ (2.67)\end{tabular} & \begin{tabular}[c]{@{}c@{}}90.03\\ (4.01)\end{tabular} & \begin{tabular}[c]{@{}c@{}}83.95\\ (1.91)\end{tabular} \\ \hline
\multirow{3}{*}{\begin{tabular}[c]{@{}c@{}}Seq2Seq-\\ Pred\end{tabular}} & $m_{\mathbf{x}}$ filtering                    & \begin{tabular}[c]{@{}c@{}}82.80\\ (4.00)\end{tabular} & \begin{tabular}[c]{@{}c@{}}53.38\\ (2.26)\end{tabular} & \multicolumn{1}{c|}{\begin{tabular}[c]{@{}c@{}}43.47\\ (1.82)\end{tabular}} & \begin{tabular}[c]{@{}c@{}}44.22\\ (1.99)\end{tabular}  & \begin{tabular}[c]{@{}c@{}}27.79\\ (1.18)\end{tabular} & \multicolumn{1}{c|}{\begin{tabular}[c]{@{}c@{}}20.44\\ (0.72)\end{tabular}} & \begin{tabular}[c]{@{}c@{}}100.00\\ (0.00)\end{tabular} & \begin{tabular}[c]{@{}c@{}}88.64\\ (9.05)\end{tabular}  & \multicolumn{1}{c|}{\begin{tabular}[c]{@{}c@{}}62.42\\ (3.85)\end{tabular}}  & \begin{tabular}[c]{@{}c@{}}89.43\\ (5.70)\end{tabular} & \begin{tabular}[c]{@{}c@{}}71.10\\ (3.15)\end{tabular} & \begin{tabular}[c]{@{}c@{}}66.78\\ (2.56)\end{tabular} \\
                                                                         & $v_{\mathbf{x}}$ filtering                     & \begin{tabular}[c]{@{}c@{}}92.06\\ (1.86)\end{tabular} & \begin{tabular}[c]{@{}c@{}}91.00\\ (1.93)\end{tabular} & \multicolumn{1}{c|}{\begin{tabular}[c]{@{}c@{}}92.32\\ (2.39)\end{tabular}} & \begin{tabular}[c]{@{}c@{}}75.78\\ (10.34)\end{tabular} & \begin{tabular}[c]{@{}c@{}}76.54\\ (4.13)\end{tabular} & \multicolumn{1}{c|}{\begin{tabular}[c]{@{}c@{}}77.96\\ (3.09)\end{tabular}} & \begin{tabular}[c]{@{}c@{}}99.98\\ (0.03)\end{tabular}  & \begin{tabular}[c]{@{}c@{}}99.99\\ (0.02)\end{tabular}  & \multicolumn{1}{c|}{\begin{tabular}[c]{@{}c@{}}99.95\\ (0.04)\end{tabular}}  & \begin{tabular}[c]{@{}c@{}}67.31\\ (6.83)\end{tabular} & \begin{tabular}[c]{@{}c@{}}77.67\\ (3.10)\end{tabular} & \begin{tabular}[c]{@{}c@{}}72.95\\ (3.19)\end{tabular} \\
                                                                         & ours                   & \begin{tabular}[c]{@{}c@{}}98.20\\ (1.11)\end{tabular} & \begin{tabular}[c]{@{}c@{}}97.79\\ (1.13)\end{tabular} & \multicolumn{1}{c|}{\begin{tabular}[c]{@{}c@{}}97.34\\ (0.95)\end{tabular}} & \begin{tabular}[c]{@{}c@{}}81.81\\ (9.05)\end{tabular}  & \begin{tabular}[c]{@{}c@{}}77.57\\ (4.18)\end{tabular} & \multicolumn{1}{c|}{\begin{tabular}[c]{@{}c@{}}81.25\\ (4.17)\end{tabular}} & \begin{tabular}[c]{@{}c@{}}100.00\\ (0.00)\end{tabular} & \begin{tabular}[c]{@{}c@{}}100.00\\ (0.00)\end{tabular} & \multicolumn{1}{c|}{\begin{tabular}[c]{@{}c@{}}100.00\\ (0.00)\end{tabular}} & \begin{tabular}[c]{@{}c@{}}95.43\\ (2.64)\end{tabular} & \begin{tabular}[c]{@{}c@{}}87.80\\ (2.98)\end{tabular} & \begin{tabular}[c]{@{}c@{}}82.55\\ (1.23)\end{tabular} \\ \hline
\end{tabular}
\end{table*}

%% file: pic_summary/lstmaepic.tex

\begin{figure}[H]
    \centering
    \includegraphics[width=\columnwidth]{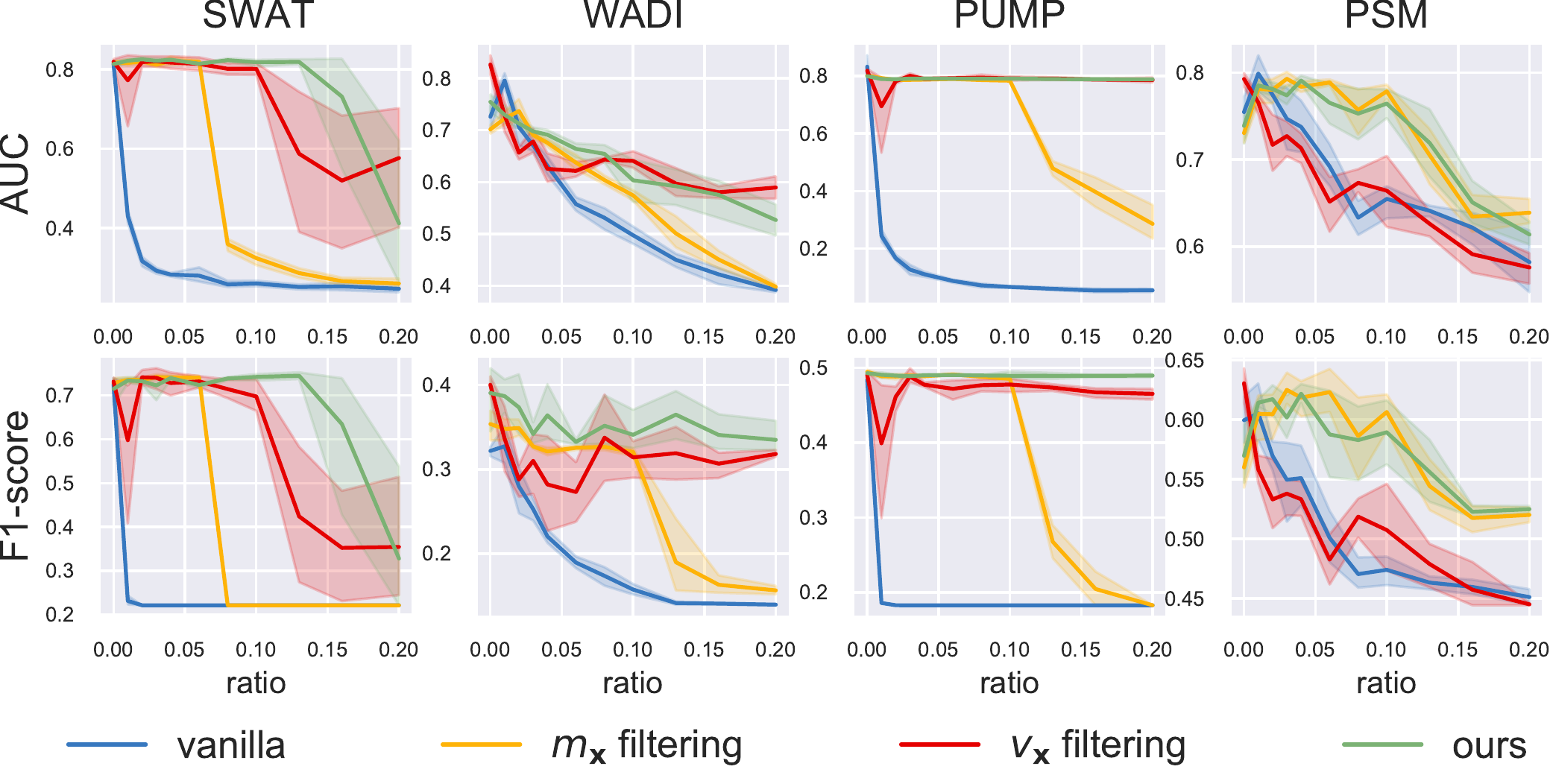}
    \caption{Performance of LSTMAE with contaminated training data under different anomaly ratios}
    \label{fig:lstmae}
\end{figure}

%% file: pic_summary/seqpic.tex
\vspace{-10pt}
\begin{figure}[H]
    \centering
    \includegraphics[width=\columnwidth]{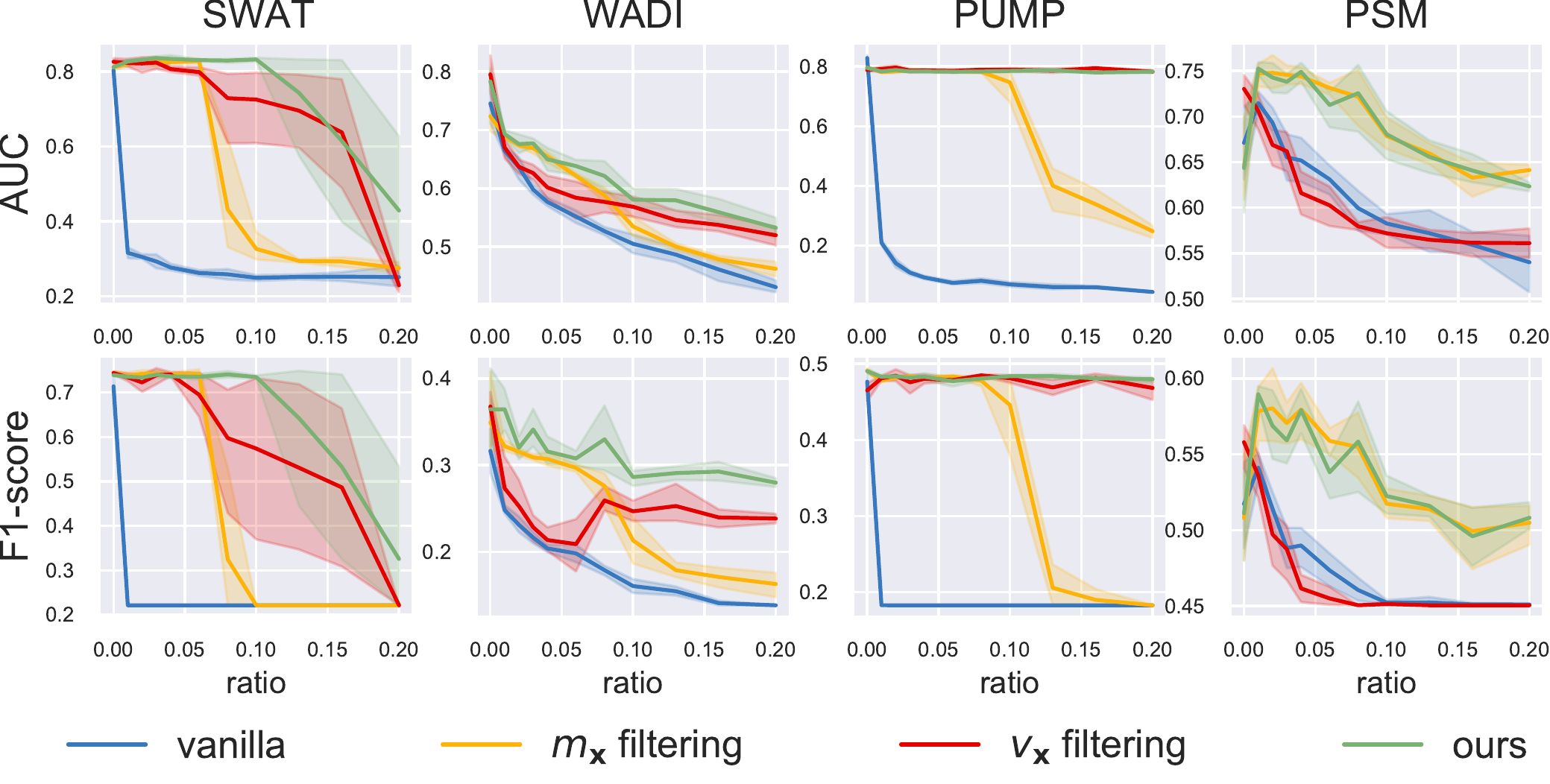}
    \caption{Performance of Seq2SeqPred with contaminated training data under different anomaly ratios}
    \label{fig:seq}
\end{figure}


%% file: pic_summary/dagmmpic.tex

\begin{figure}[H]
    \centering
    \includegraphics[width=\columnwidth]{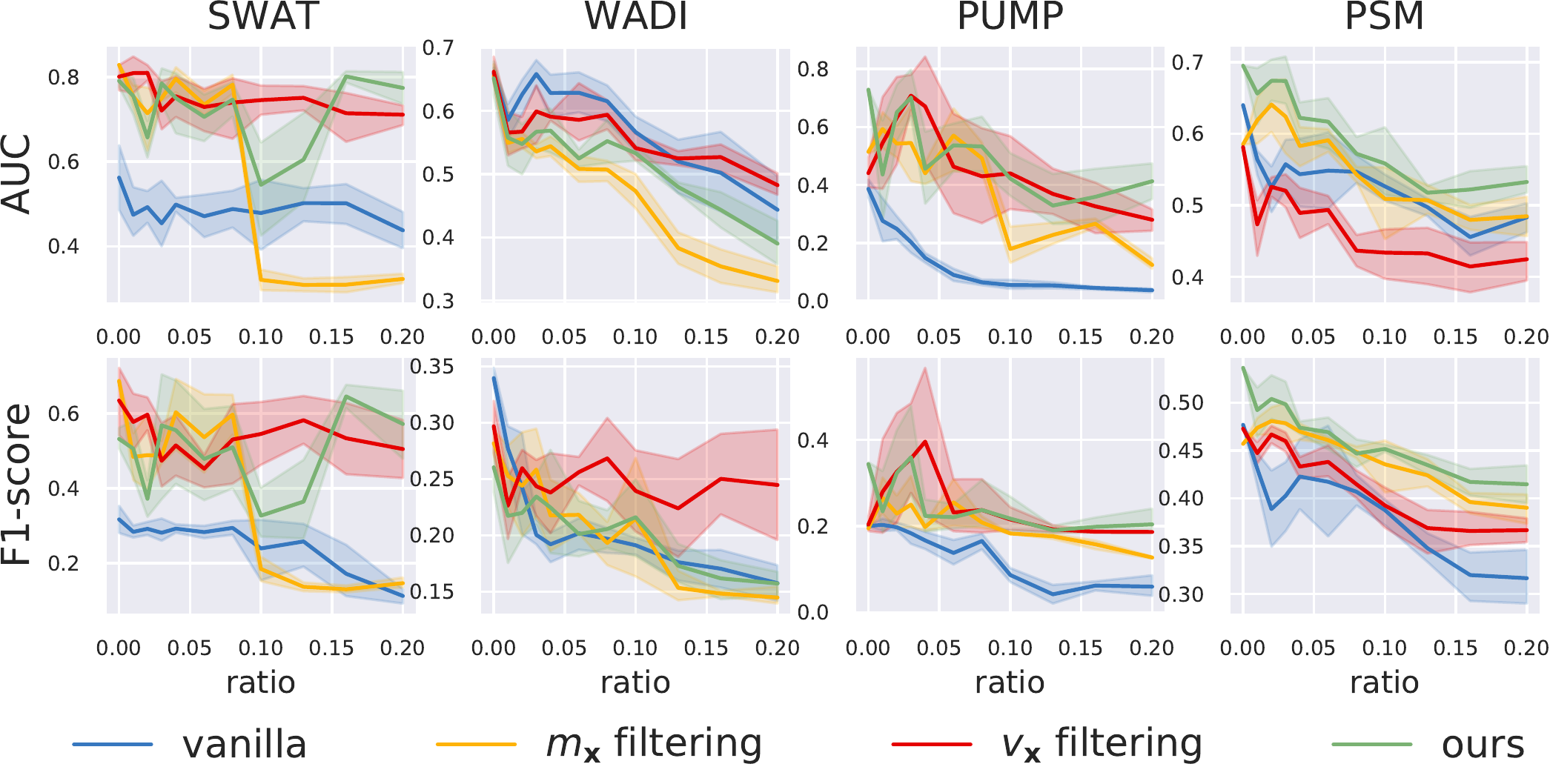}
    \caption{Performance of DAGMM with contaminated training data under different anomaly ratios}
    \label{fig:dagmm}
\end{figure}

%% file: pic_summary/deepsvddpic.tex
\vspace{-10pt}
\begin{figure}[H]
    \centering
    \includegraphics[width=\columnwidth]{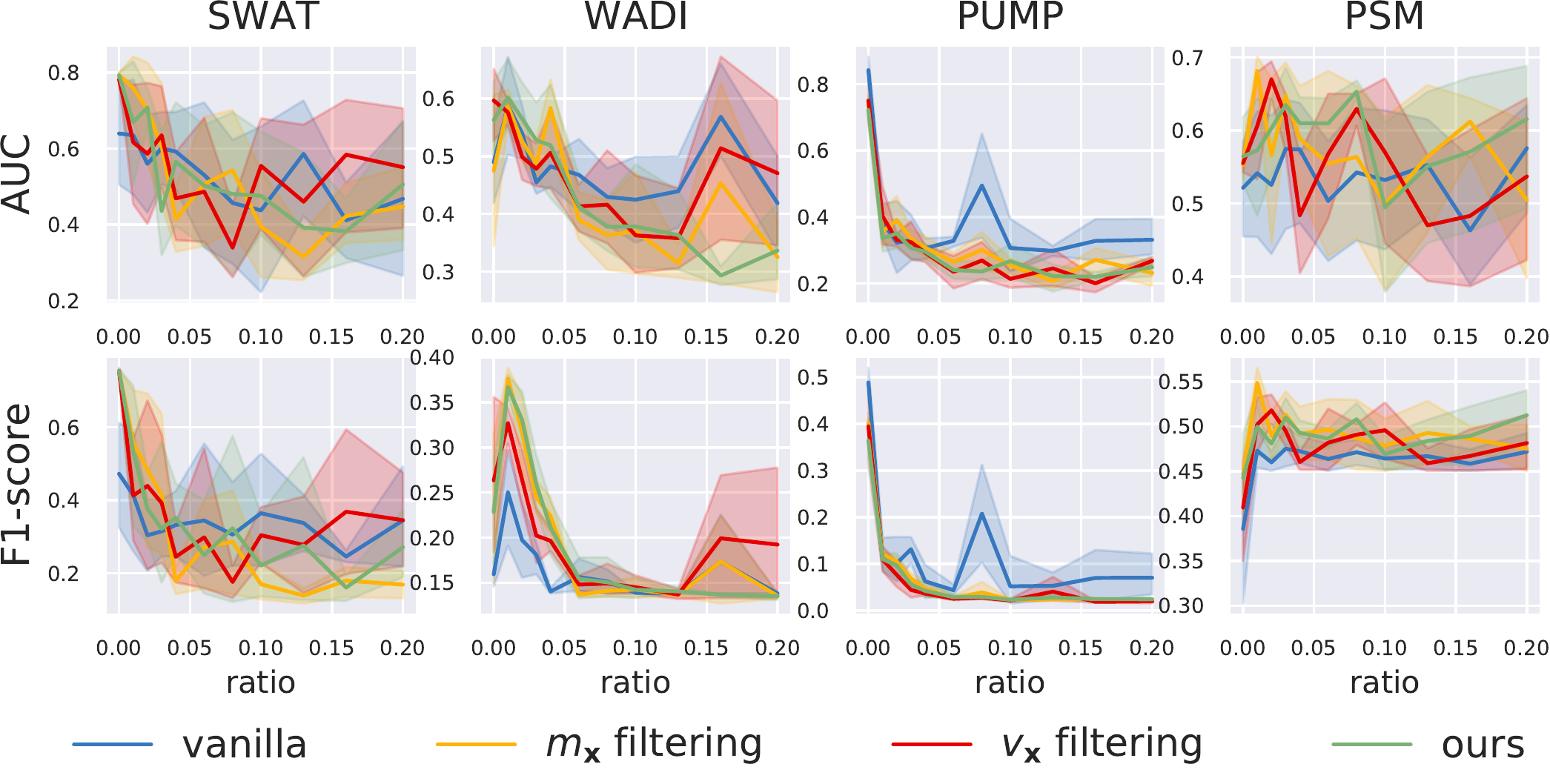}
    \caption{Performance of DeepSVDD with contaminated training data under different anomaly ratios}
    \label{fig:deepsvdd}
\end{figure}
